\documentclass[10pt,twocolumn,letterpaper]{article}

\usepackage{cvpr}              

\newcommand{\red}[1]{{\color{red}#1}}

\definecolor{cvprblue}{rgb}{0.21,0.49,0.74}
\usepackage[pagebackref,breaklinks,colorlinks,allcolors=cvprblue]{hyperref}
\usepackage{multirow}
\usepackage{amsmath}
\usepackage{amsfonts}
\usepackage{geometry}
\usepackage{booktabs}
\usepackage{array}
\usepackage{colortbl}
\usepackage{pifont}
\usepackage{fontawesome5}
\usepackage{bm}

\definecolor{lightblue}{RGB}{100,160,255}   
\definecolor{lightred}{RGB}{255,120,120}    

\usepackage{booktabs}      
\usepackage{makecell}      
\def\confName{CVPR}
\def\confYear{2026}

\usepackage[table]{xcolor} 
\usepackage[notransparent]{svg}
\NewDocumentCommand{\inlineimage}{O{0.5} m}{%
  \raisebox{-0.2\baselineskip}{\includegraphics[height=#1\baselineskip]{#2}}\hspace{-3pt}
}

\definecolor{mindM}{HTML}{1F82C0}  
\definecolor{mindI}{HTML}{1CBF91}  
\definecolor{mindN}{HTML}{39C46E}  
\definecolor{mindD}{HTML}{149C7E}  

\definecolor{my_green}{RGB}{51,102,0}
\definecolor{my_red}{RGB}{204, 0, 0}
\renewcommand{\checkmark}{\textcolor{my_green}{\ding{51}}} 
\newcommand{\crossmark}{\textcolor{my_red}{\ding{55}}} 

\newcommand{\logo}{\raisebox{0\height}{\inlineimage[1.2]{figure/logo.pdf}}}

\newcommand{\ours}{\textcolor{mindM}{M}%
\textcolor{mindI}{I}%
\textcolor{mindN}{N}%
\textcolor{mindD}{D}\xspace
}

\title{\logo \space \ours: Benchmarking Memory Consistency and Action Control \\ in World Models}

\author{Yixuan Ye$^{1*}$, Xuanyu Lu$^{1*}$, Yuxin Jiang$^{2*}$, Yuchao Gu$^{2}$, Rui Zhao$^{2}$, Qiwei Liang$^{3}$, Jiachun Pan$^{2}$, \\Fengda Zhang$^{4}$, Weijia Wu$^{2 \dagger}$, Alex Jinpeng Wang $^{1 \dagger}$\\
$^1$ CSU-JPG, Central South University $^2$ National University of Singapore\\
$^3$ Hong Kong University of Science and Technology (Guangzhou) $^4$ Nanyang Technological University\\
Project Page: \url{https://csu-jpg.github.io/MIND.github.io/}
}

\begin{document}

\makeatletter
\let\@oldmaketitle\@maketitle
\renewcommand{\@maketitle}{\@oldmaketitle
 \centering
    \includegraphics[width=0.99\textwidth]{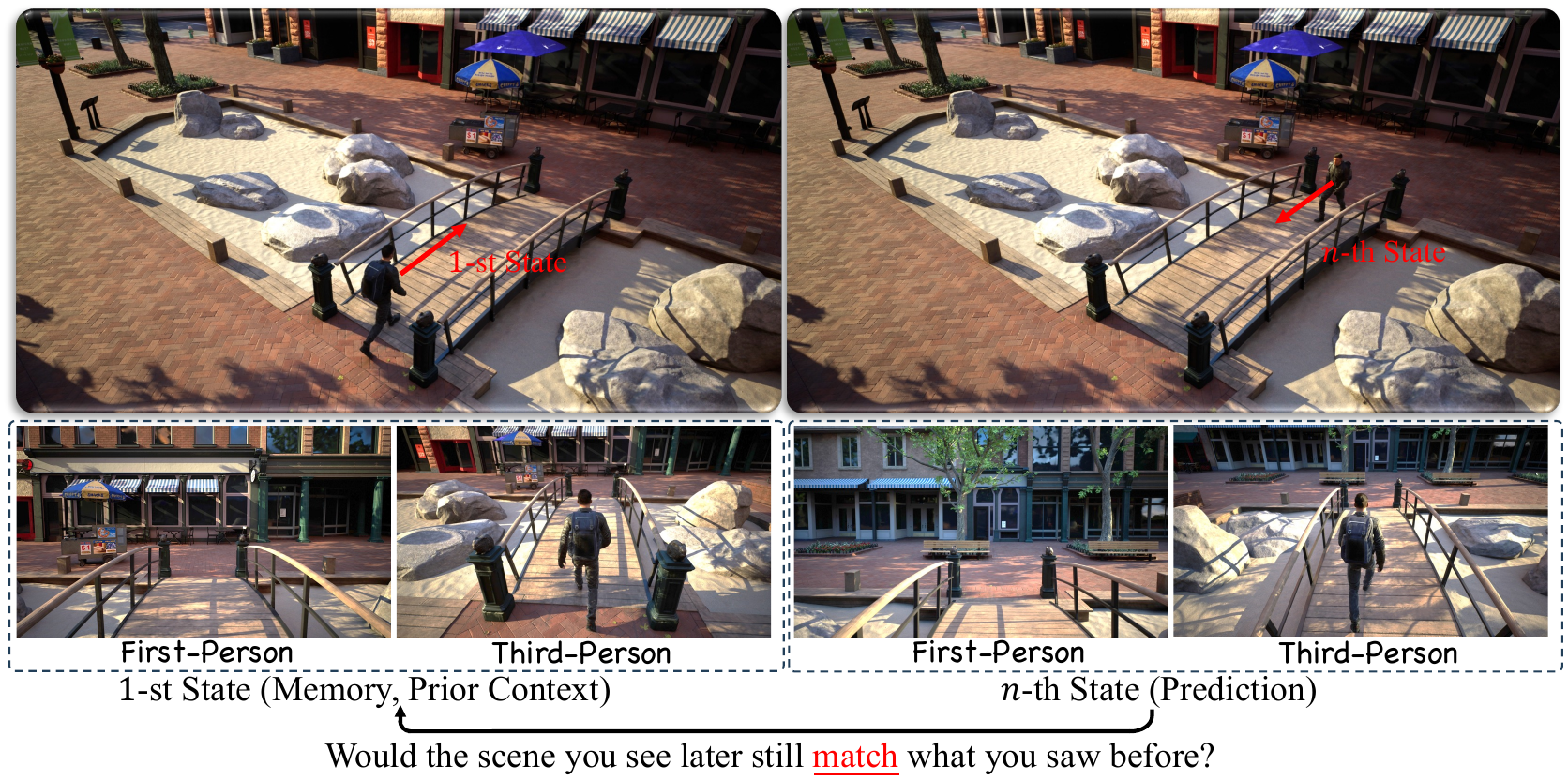}
    \vspace{-0.4cm}
    \captionof{figure}{\textbf{Evaluation for Memory Consistency and Action Control with \ours}. 
    The first open-domain closed-loop revisited benchmark at $1080$p/$24$ FPS for evaluating world models from both first-person and third-person perspectives.}
    \label{fig-motivation}
  \bigskip}
\makeatother
\maketitle

\begin{abstract}

World models aim to understand, remember, and predict dynamic visual environments, yet a unified benchmark for evaluating their fundamental abilities remains lacking.
To address this gap, we introduce \ours, the first open-domain closed-loop revisited benchmark for evaluating \textbf{M}emory cons\textbf{I}stency and action co\textbf{N}trol in worl\textbf{D} models.
\ours\ contains 250 high-quality videos at 1080p and 24 FPS, including 100 (first-person) + 100 (third-person) video clips under a shared action space and 25 + 25 clips across varied action spaces covering eight diverse scenes.
We design an efficient evaluation framework to measure two core abilities: \textbf{memory consistency} and \textbf{action control}, capturing temporal stability and contextual coherence across viewpoints.
{\let\thefootnote\relax\footnotetext{$^*$ Equal contribution. $^\dagger$ Corresponding author.}}
Furthermore, we design various action spaces, including different character movement speeds and camera rotation angles, to evaluate the \textit{action generalization} capability across different action spaces under shared scenes.
To facilitate future performance benchmarking on \ours, we introduce \textbf{MIND-World}, a novel interactive Video-to-World baseline.
Extensive experiments demonstrate the completeness of \ours\ and reveal key challenges in current world models, including the difficulty of maintaining long-term memory consistency and generalizing across action spaces.

\end{abstract}    
\section{Introduction}
\label{sec:intro}

\begin{table*}[t]
    \centering
      \footnotesize
      \caption{\textbf{Comparison of World Model Benchmarks.} 
      `Avg.' denotes the average number of frames used for 
        \textcolor{lightblue}{memory context} and \textcolor{lightred}{predicted segment} in each benchmark.
      `$1$st-P.' and `$3$rd-P.' refer to first person and third person perspectives, respectively.
      $\emptyset$ denotes the benchmarks without action-based generation (\textit{e.g.,} text–video gen).
      `CharPos.' refer to the character position.
      \ours is the first open-domain closed-loop revisited benchmark for evaluating video consistency across both first- and third-person perspectives.
    }
    \vspace{-2mm}
    \label{tab:comparsion_bench}
\begin{tabular}{l|c|cc|cc|cp{0.5\columnwidth}}
    \multirow{2}{*}{\textbf{Benchmark}} & \multirow{2}{*}{\textbf{CharPos.}} & \multicolumn{2}{c|}{\textbf{Fixed Act.}~(Avg.)}   & \multicolumn{2}{c|}{\textbf{Generalized Act.}~(Avg.)}   & \multirow{2}{*}{\textbf{Res./FPS}}    &  \multirow{2}{*}{\textbf{Scenario (image\faImage/video\faVideo)}} 
    \\
    \cline{3-6}
    
     &  & $1$st-P.     & $3$rd-P.  & $1$st-P.   & $3$rd-P.   &      & 
     \\
    \hline

    WorldSimBench~\cite{qin2025worldsimbench} & \crossmark & \textcolor{lightblue}{1} / \textcolor{lightred}{-} & $\emptyset$ & $\emptyset$ &  \textcolor{lightblue}{1} / \textcolor{lightred}{-} & -/- & \faImage: Minecraft, Driving... \\


    WorldModelBench~\cite{Li2025WorldModelBench} & \crossmark & \textcolor{lightblue}{1} / \textcolor{lightred}{-} & $\emptyset$ & $\emptyset$ & $\emptyset$ & -/- & \faImage: Humans, Natural... \\


    WorldScore~\cite{duan2025worldscore} & \crossmark & \textcolor{lightblue}{1} / \textcolor{lightred}{-} & $\emptyset$ & $\emptyset$ & $\emptyset$ & -/- & \faImage: Dining , Passageways... \\


    World-in-World~\cite{zhang2025world} & \crossmark & $\emptyset$ & $\emptyset$ & $\emptyset$ & $\emptyset$ & 576p/- & \faImage: Interior environment... \\
    

    GameWorld~\cite{zhang2025matrix} & \crossmark & \textcolor{lightblue}{1} / \textcolor{lightred}{-} & $\emptyset$ & $\emptyset$ & $\emptyset$ & 720p/- & \faImage: Minecraft \\
    

    Lian et al. \cite{lian2025toward}    & \checkmark & \textcolor{lightblue}{65} / \textcolor{lightred}{436} & $\emptyset$ & $\emptyset$ & $\emptyset$ & 360p/20 & \faImage,\faVideo: Minecraft \\
    
    \hline
    \ours ~(Ours)   & \checkmark & \textcolor{lightblue}{1.1k}/\textcolor{lightred}{3.4k} & \textcolor{lightblue}{1.2k}/\textcolor{lightred}{3.6k} & \textcolor{lightblue}{1.3k}/\textcolor{lightred}{3.8k} & \textcolor{lightblue}{1.2k}/\textcolor{lightred}{3.7k} & 1080p/24
     & \faImage,\faVideo: Landscape, SciFi,\\& & & & & & &Stylized, Ancient, Urban, \\& & & & & & &Industrial, Interior, Aquatic \\
    
    \end{tabular}

\end{table*}

Recent advances in video generation technology have significantly improved the creation of high-fidelity, realistic content, laying a solid foundation for developing sophisticated world models~\cite{he2025matrix,ye2025yan,yu2025gamefactory,cui2025emu35nativemultimodalmodels,mao2025yume}.
These models have accelerated advancements across diverse domains, including autonomous driving~\cite{li2025drivevla,mousakhan2025orbis,yang2025raw2drive,li2025omninwmomniscientdrivingnavigation,zhang2025epona}, embodied intelligence~\cite{cen2025worldvla,lv2025f1,bruce2024genie,panteam2025panworldmodelgeneral,chi2025wow}, and interactive game environments~\citep{che2024gamegen,valevski2024diffusion,xiao2025worldmem,li2025hunyuangamecrafthighdynamicinteractivegame,ye2025yan}, by enabling the generation of complex, diverse, and controllable virtual worlds.
Despite these advances, building a reliable world model remains challenging.
Beyond visual realism, such models must maintain long-term memory consistency and exhibit accurate action control and robust action generalization across diverse scenarios.
Yet, current evaluations mainly focus on visual quality or physical realism, overlooking these essential aspects.
Consequently, the field still lacks a comprehensive benchmark to systematically assess memory consistency and action controllability in open-domain environments.

%

%
Existing benchmarks primarily focus on evaluating the \textit{quality} and \textit{realism} of generated videos, often limited to \textit{first-person} perspective data collected within a single action space.
For instance, WorldScore \citep{duan2025worldscore} decomposes scene generation into specific camera motion trajectories to assess video quality, while WorldModelBench \cite{Li2025WorldModelBench} evaluates adherence to physical laws to measure world modeling capabilities in application-driven domains.
Although Lian et al.~\cite{lian2025toward} introduced a world model memory benchmark, it is limited to Minecraft scenes, lacks open-domain diversity, and depends on loop-based agent data that poorly reflects human behavior.
Furthermore, the existing world model benchmark predominantly features first-person perspectives \cite{duan2025worldscore,lian2025toward}, making it challenging to evaluate the ability of world models to simulate motion and poses.
In summary, as shown in Table~\ref{tab:comparsion_bench}, existing benchmarks focus mainly on \textit{first-person settings} and \textit{image-level evaluation}, lacking \textit{memory consistency} assessment and scene diversity. 
Establishing a comprehensive world model benchmark remains an open and unresolved challenge.

We present \ours, the first closed-loop revisited open-domain benchmark for evaluating memory consistency and action control from both first-person and third-person perspectives across diverse scenarios.
\ours focuses on two key abilities of world models: \textbf{1) Memory consistency} refers to the ability of model to maintain coherent spatial layouts, object identities, and scene attributes over long temporal contexts, ensuring that generated frames remain consistent with past observations.
2) \textbf{Action control} measures how accurately the model executes given control inputs and generalizes these dynamics to new motion ranges or unseen action spaces, reflecting its capacity for precise and adaptable interaction within dynamic environments.
Furthermore, the provided videos include frame-level aligned actions, character and camera positions, and image labels, collected from multiple volunteers to capture diverse human behaviors.
The dataset contains $250$ high-quality $1080$p / $24$ FPS, frame-level action-aligned videos spanning eight major scene categories, enabling comprehensive evaluation of world models.
%

%
%

To summarize, the contributions of this paper are: 
\begin{itemize}[leftmargin=*]
    \item \textbf{Open-Domain Benchmark for World Models.} 
    We introduce \ours, the first closed-loop revisited open-domain benchmark at $1080$p / $24$ FPS for evaluating world models from both \textit{first-person} and \textit{third-person} perspectives.

    \item \textbf{Evaluation for Memory Consistency and Action Control.} 
    We design an efficient framework to assess \textit{memory consistency} and \textit{action control}, capturing temporal stability and contextual coherence across viewpoints.

    \item \textbf{Evaluation for Cross-Action Space Generalization.} 
    We design various action spaces, including different character movement speeds and camera rotation angles, to evaluate the \textit{action space generalization} capability.

    \item \textbf{The novel Video-to-World baseline, MIND-World.}
    Extensive experiments demonstrate the completeness of \ours\ and expose key challenges in current world models, such as limited long-term memory consistency and limited generalization across action spaces.
\end{itemize}


\section{Related Work}
\label{sec:Related}
\setlength{\parindent}{0pt}

\begin{figure*}[t]
    \centering
    \includegraphics[width=0.96\linewidth]{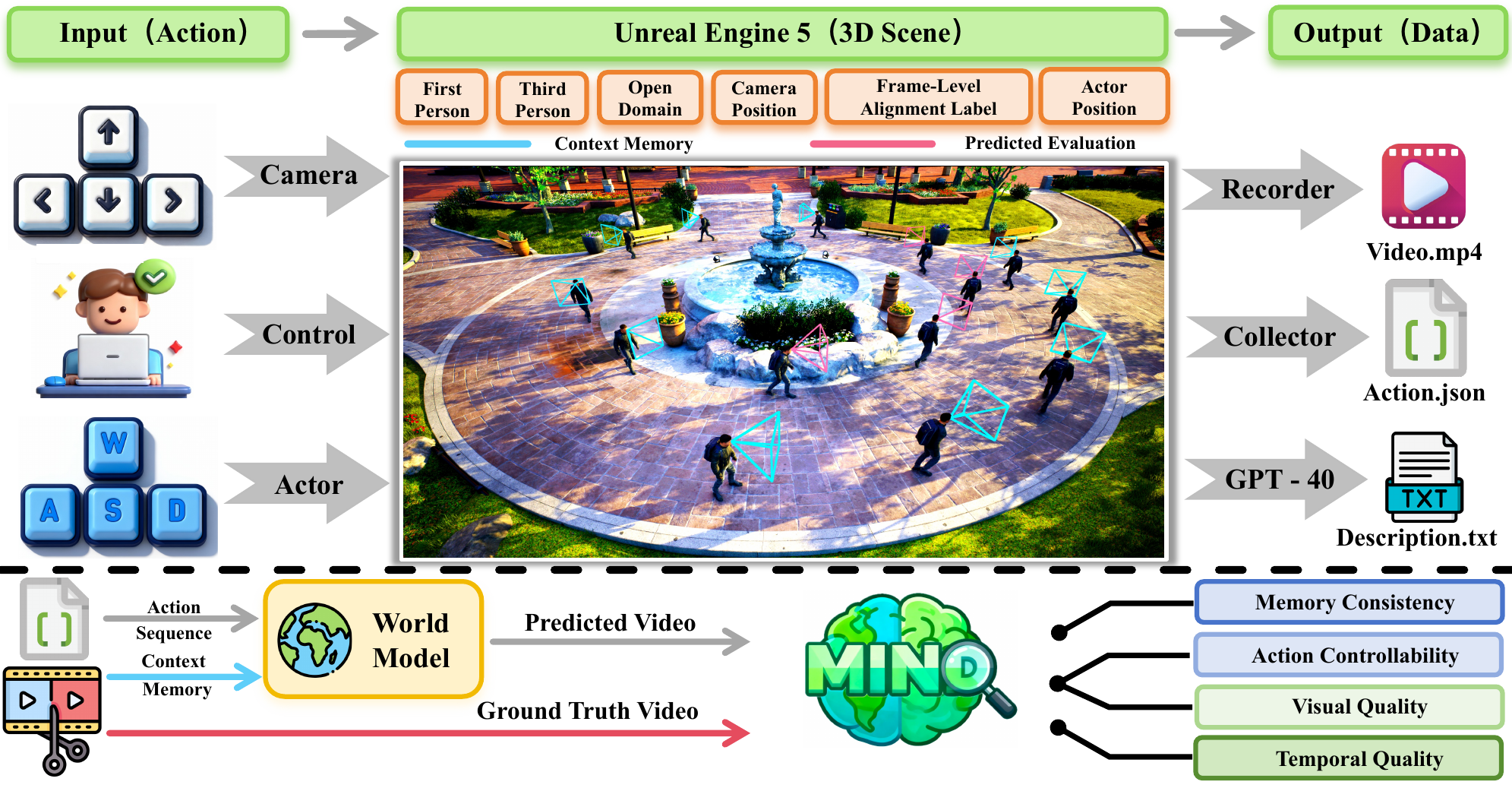}
    \vspace{-0.3cm}
    \caption{\textbf{Overview of the \ours}. 
    We build and collect the first open-domain closed-loop revisited benchmark using Unreal Engine 5, supporting both first-person and third-person perspectives with $1080$ p resolution at $24$ FPS.}
    \label{fig-data-pipeline}
\end{figure*}

\subsection{Video Generation}

Recently, video generation models represented by SVD~\citep{svd}, Hunyuanvideo~\citep{kong2024hunyuanvideo}, Wan~\citep{wan2025wan} and Sora 2~\citep{openai_sora2_2025} have achieved remarkable breakthroughs. While significantly enhancing the visual realism, temporal consistency and controllability of generated videos, these models have extended their generation capabilities to long-sequence and physically plausible scenarios. Benchmarks including VBench~\citep{huang2024vbench} and VBench-2.0~\citep{zheng2025vbench} have established dedicated evaluation systems for video generation models, covering key dimensions such as human fidelity, physical plausibility and commonsense consistency.


\subsection{World Model}
Recent advances in world models have broken down the technical barriers between visual generation and embodied simulation, enabling agents or users to interact in temporally consistent virtual environments.
Unlike traditional text-to-video models, world models emphasize long-term memory consistency \citep{wu2025videoworldmodelslongterm,huang2025memoryforcing,wu2026infiniteworld,gu2025long,yu2025contextmemory,li2025vmem,zhao2025spatia}, action-conditioned controlled generation \citep{yu2025gamefactory,ye2025yan,tang2025hunyuan,hong2025relic,gao2025adaworld} and real-time response \citep{huang2025self,bruce2024genie,team2026advancing,he2025matrix,sun2025worldplay}, evolving into three core research directions.
To ensure long-term memory consistency, mainstream existing studies adopt three strategies: pose frame retrieval, context memory compression, and explicit 3D memory representation. 
Specifically, \texttt{CAM} \cite{yu2025contextmemory} retrieves context frames based on the field-of-view coverage of pose perspectives, \texttt{Infinite-World} \cite{wu2026infiniteworld} designs a hierarchical pose-free memory compression module to autonomously anchor generated content to distant historical information, and \texttt{SPMem} \cite{wu2025videoworldmodelslongterm} achieves explicit 3D memory representation by virtue of geometrically anchored long-term spatial memory.
For the optimization of action-conditioned controlled generation, \texttt{GameFactory} \cite{yu2025gamefactory} proposes a multi-stage training strategy integrated with domain adapters, which decouples game style learning from action control to realize scene-generalizable action control and \texttt{AdaWorld} \cite{gao2025adaworld} embeds action information into the pre-training process and extracts implicit actions from videos via self-supervision, thus enabling novel action learning under limited conditions.
Real-time interaction is a core characteristic of this field, and relevant training paradigms lay a foundation for real-time streaming generation of diffusion-based world models. For instance, \texttt{Diffusion-Forcing} \cite{chen2024diffusion} trains diffusion models to denoise token sets with independent per-token noise levels and \texttt{Self-Forcing} \citep{huang2025self} performs autoregressive inference with KV caching during training, conditioning the generation of each frame on the model’s own prior outputs. 
Together, these advances mark a shift from static video synthesis to interactive, temporally consistent world models.

\subsection{Evaluation for World Model}
The rapid rise of world models has spurred new benchmarks, yet most primarily emphasize scene quality or physical plausibility.
WorldScore~\citep{duan2025worldscore} standardizes camera-trajectory layouts to rate generated video quality.
WorldModelBench~\citep{Li2025WorldModelBench} targets adherence to physical laws in application-driven settings.
And WorldSimBench~\citep{qin2025worldsimbench} assesses visual realism.
However, these efforts under represent two core abilities of world models: long-context \textbf{memory consistency} and \textbf{action-space generalization} across varied controls.
In contrast, \ours introduces the first open-domain closed-loop revisited benchmark at 1080p and 24\,FPS from both first-person and third-person views, providing unified, efficient protocols to evaluate memory consistency and action control.

\section{\ours Benchmark}
\label{sec:Benchmark}


\subsection{Video Source and Environment Settings}
\label{VideoSource}

To comprehensively evaluate world models across diverse interactive contexts, we construct a large-scale video corpus rendered within Unreal Engine 5.
As shown in Figure \ref{fig-scene}, the benchmark spans $8$ categories covering over $40$ open-domain environments, designed to reflect a wide spectrum of visual and physical dynamics.
These include natural (\textit{e.g.,} forest, desert, mountain, ocean), urban (\textit{e.g.,} downtown, residential, industrial), indoor, vehicle, sci-fi, stylized, fantasy, and abstract scenes.
%
%
As shown in Figure \ref{fig-data-pipeline} , we construct a systematic data generation pipeline and recruit multiple volunteers to perform both scripted and free-form actions within these environments.
%
%
We collect $250$ frame-aligned videos at $1080$ p / $24$ FPS. Among them, $200$ videos ($100$ first-person and $100$ third-person, evenly split for training and testing) share the same action space, while the remaining $50$ videos ($25$ per perspective) feature distinct action spaces, providing high-quality and controllable ground truth for evaluation.

%
%
%
%

\subsection{Basic Actions Modeling}
\label{ActionSpaceDesign}
In this section, we define a \textbf{basic action set} for modeling both \textbf{agent translation} and \textbf{camera rotation}, which are essential for evaluating action control and scene consistency in world models.

\textbf{Action Space Definition.} We define the action space \( \mathcal{A} \) as follows:
\[
\mathcal{A} = \{W, A, S, D, \uparrow, \downarrow, \leftarrow, \rightarrow\},
\]
where:
\begin{itemize}
    \item \( W, A, S, D \) correspond to \textbf{forward}, \textbf{left}, \textbf{backward}, and \textbf{right} movement,
    \item \( \uparrow, \downarrow, \leftarrow, \rightarrow \) correspond to \textbf{camera pitch} and \textbf{yaw} rotations.
\end{itemize}

\begin{figure}[t]
    \centering
    \includegraphics[width=0.99\linewidth]{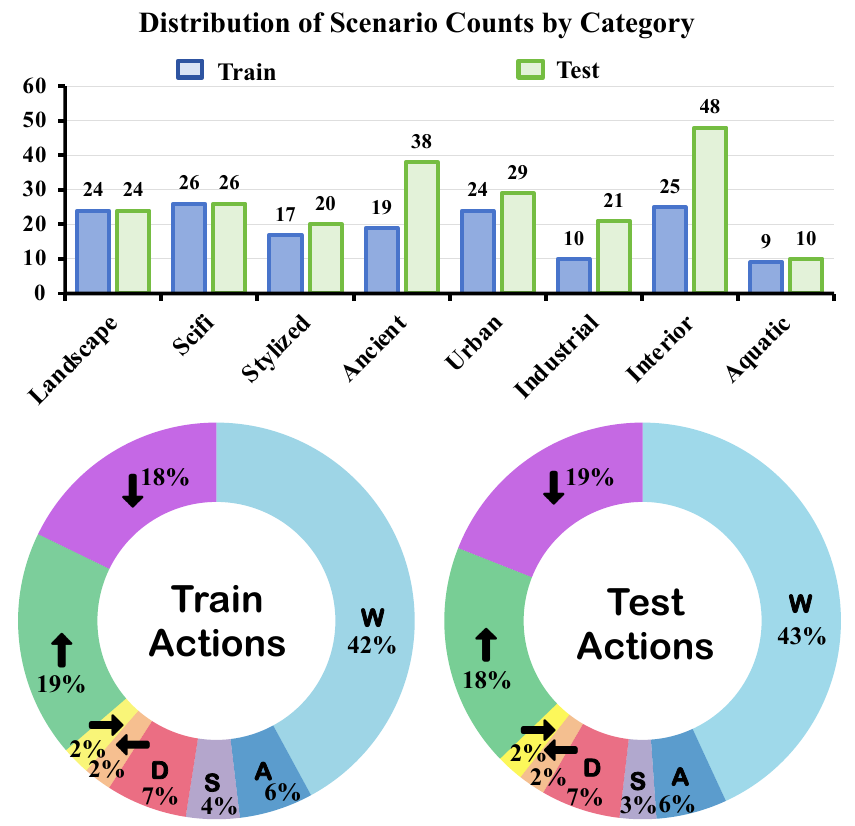}
    \vspace{-0.3cm}
    \caption{\textbf{Distribution for Scene Categories and Action Space in \ours.} 
    \ours supports open-domain scenarios with diverse and well-balanced action spaces. }
    \label{fig-scene_number}
\end{figure}

\textbf{Translational Motion}. For translational actions, the position of agent \( \mathbf{p}_t = [x_t, y_t, z_t]^\top \) is updated based on the selected movement direction:
\[
\mathbf{p}_{t+1} = \mathbf{p}_t + \Delta_p \cdot \mathbf{v}_a,
\]
where \( \mathbf{v}_a \) is the direction of movement corresponding to the action (e.g., \( \mathbf{v}_W = [0,0,1]^\top \) for forward) and \( \Delta_p \) is the step size.

\textbf{Rotational (Camera) Motion.} For camera rotation, the orientation \( \mathbf{r}_t = [\theta_t, \phi_t]^\top \) is updated by a small angular increment \( \Delta_r \):
\[
\mathbf{r}_{t+1} = \mathbf{r}_t + \Delta_r \cdot \mathbf{u}_a,
\]
where \( \mathbf{u}_a \) corresponds to the direction of camera rotation (e.g., \( \mathbf{u}_{\uparrow} = [0, +1]^\top \) for pitch up).

\begin{figure}[t]
    \centering
    \includegraphics[width=0.99\linewidth]{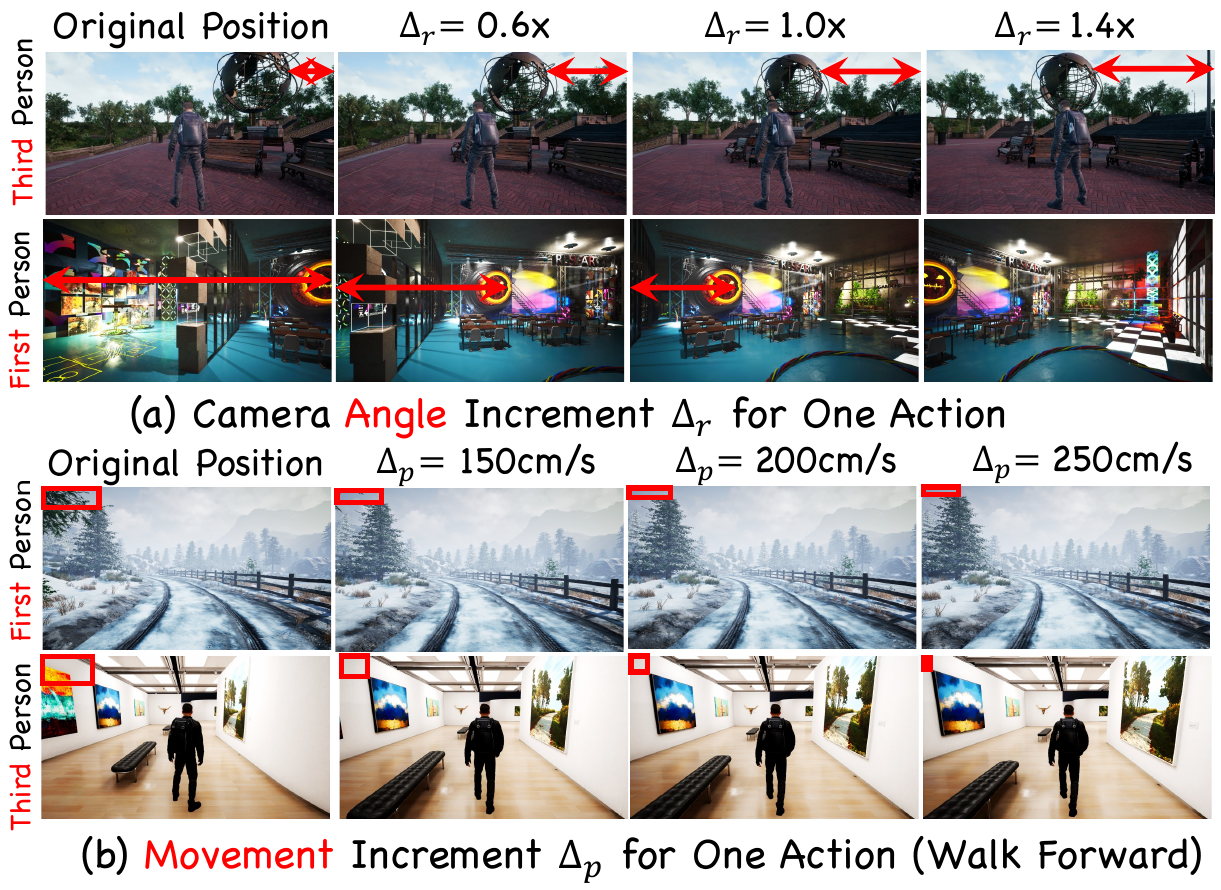}
    \vspace{-0.5cm}
    \caption{\textbf{Action Generalization from \ours}. 
    Different generalization settings for \( \Delta_p \) (movement increment) and \( \Delta_r \) (camera angle increment) are derived from both first-person and third-person perspectives. Each image is captured after the action has been executed for 24 frames.}
    \label{fig-action-gen}
\end{figure}

\subsection{Action Space Generalization}
In the action modeling framework, the values of \( \Delta_p \) (for translational motion) and \( \Delta_r \) (for rotational motion) are not fixed, but can be generalized to accommodate a range of action scales.
%
%
The action set can be customized to represent different motion scales, as shown in Figure~\ref{fig-action-gen}.
For example, an action with a \textbf{0.7-degree} rotation and \textbf{150 units} of translation (\( \Delta_r = 0.7^\circ, \Delta_p = 150 \)) allows for precise control. 
Larger movements, such as a \textbf{1.4-degree} rotation with \textbf{280 units} of translation (\( \Delta_r = 1.4^\circ, \Delta_p = 280 \)), represent broader actions.
Conversely, smaller steps like \textbf{0.4 degrees} of rotation with \textbf{100 units} of translation (\( \Delta_r = 0.4^\circ, \Delta_p = 100 \)) enable more subtle adjustments, useful for tasks requiring high precision.
This flexibility in \( \Delta_p \) and \( \Delta_r \) allows the system to adapt to varying levels of control and task requirements.
%


%
%
Action generalization enhances the flexibility of model and realism across diverse scenarios.
Thus, world models must adapt to varied action spaces.
To assess this adaptability, we collect high-quality, frame-aligned videos from different action spaces within the same scene.
Specifically, we configure five combinations of character movement speeds \( \Delta_p \) and camera rotation speeds \( \Delta_r \) to generate datasets with diverse action spaces.
The setting includes a total of $25$ first-person and $25$ third-person clips, thereby comprehensively and systematically assessing the generalization capability of world models.

\subsection{Temporal and Memory Consistency}
To evaluate the ability of world models to maintain memory and temporal consistency over time, we introduce a memory revisit strategy, as illustrated in Figure~\ref{fig-data-pipeline}.
In our setup, a human operator performs predefined actions (\( W, A, S, D, \uparrow, \downarrow, \leftarrow, \rightarrow \)) within a 3D Unreal Engine~5 environment. 
The resulting first-person and third-person videos are frame-aligned with action logs and used as ground-truth supervision.

\textbf{Memory Setup.} We define a \textit{memory segment} as a observed video sequence \( \mathcal{M} = \{f_1, f_2, \dots, f_T\} \), where each frame \( f_t \) encodes both visual appearance and scene layout.
The memory provides contextual grounding that the model must retain when generating subsequent predictions.
After observing \( \mathcal{M} \), the model receives an action sequence \( \mathcal{A} = \{a_{T+1}, \dots, a_{T+k}\} \) and is required to predict the future video frames \( \hat{\mathcal{V}} = \{\hat{f}_{T+1}, \dots, \hat{f}_{T+k}\} \).

\textbf{Consistency Objective.} The model is evaluated on whether the predicted frames \( \hat{f}_{T+i} \) remain temporally and spatially consistent with the memorized scene. This includes:
\begin{itemize}
    \item \textbf{Memory Consistency:} previously observed objects, layouts, and textures should remain unchanged when revisited through new actions (e.g., returning to the same location should reproduce the same scene appearance);
    \item \textbf{Temporal Consistency:} predicted frames should exhibit smooth transitions and coherent dynamics over time, avoiding flickering or sudden structural changes.
\end{itemize}
Formally, given a revisiting trajectory \( \mathcal{A}_{\text{loop}} \) that leads back to a previously seen state, the consistency error can be defined as:
\[
\mathcal{L}_{\text{mem}} = \| \hat{f}_{t} - f_{t'} \|_2^2,
\]
where \( f_{t'} \) corresponds to the ground-truth frame at the revisited scene.

\subsection{Evaluation}
\label{Evaluation}
\begin{figure}[t]
    \centering
    \includegraphics[width=1.0\linewidth]{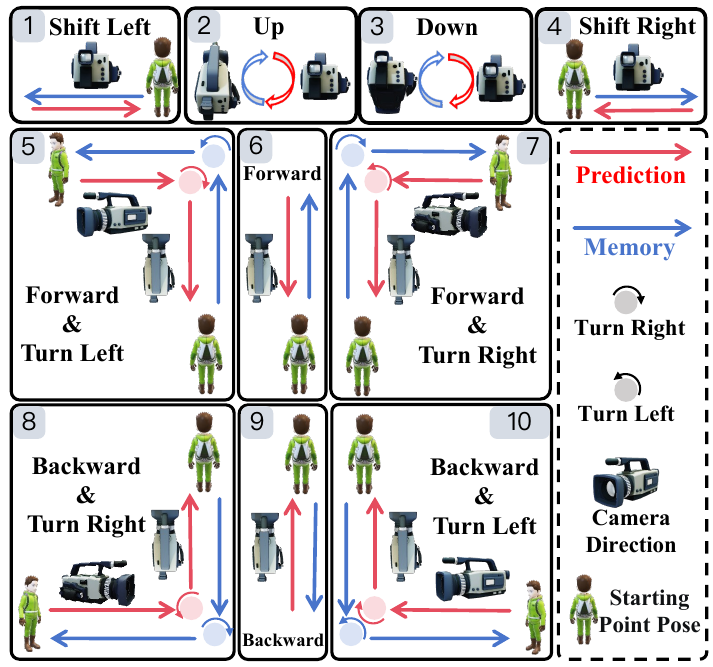}
    \vspace{-5mm}
    \caption{\textbf{The 10 Symmetric Motion Paths}. 
    The blue line represents the original path, and the red line represents the corresponding mirrored path. Each action lasts $24$ frames.}
    \label{fig-mirror-path}
\end{figure}

\begin{figure*}[t]
    \centering
    \includegraphics[width=0.99\linewidth]{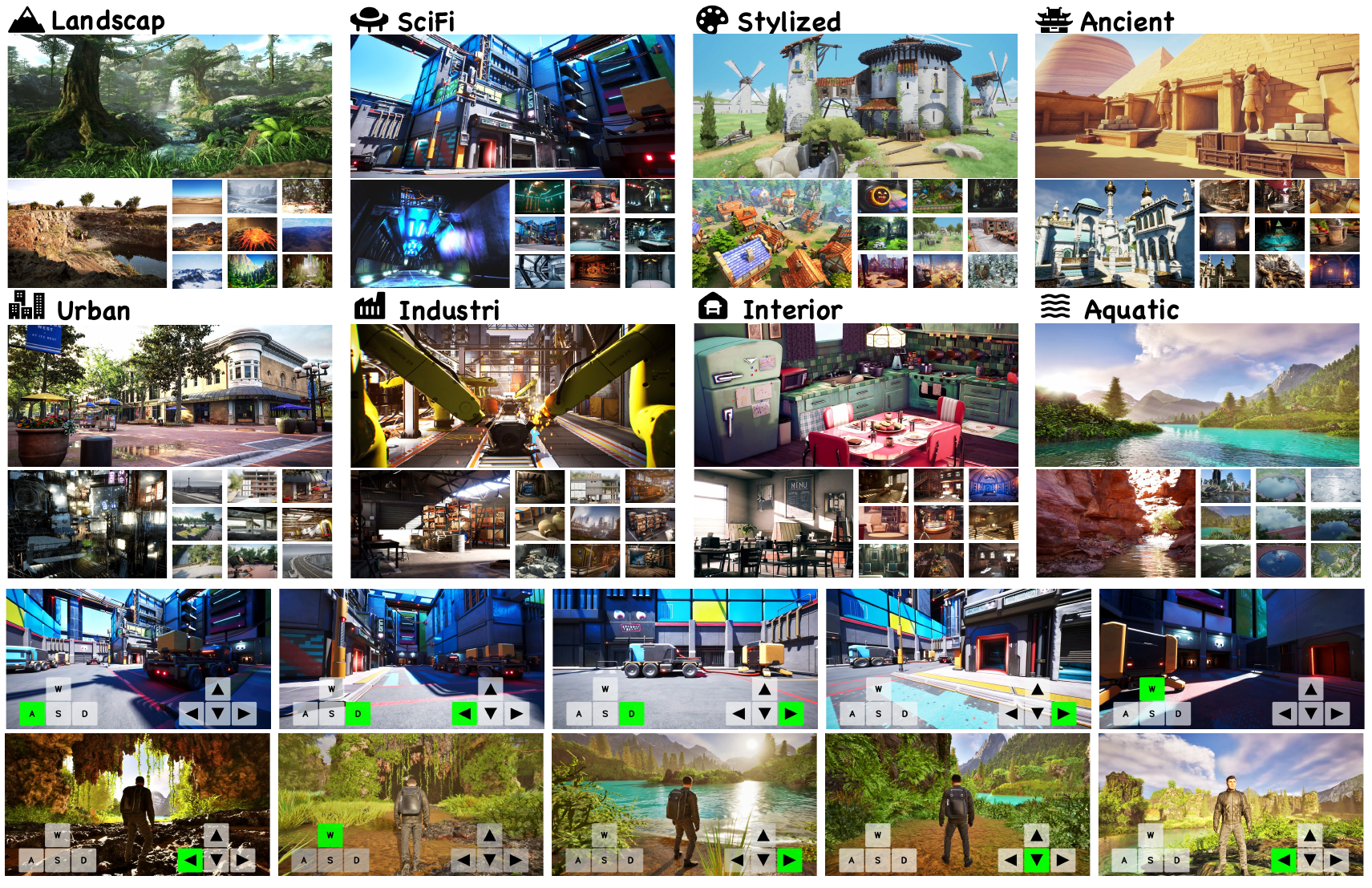}
    \vspace{-0.3cm}
    \caption{\textbf{Eight Scene Categories and Action Visualization in \ours.} 
    Each category covers multiple representative environments designed to evaluate action-following controllability and history consistency in world models. }
    \label{fig-scene}
\end{figure*}


\textbf{Long Context Memory Consistency.}
\textit{Long context memory} evaluates the ability of world model to reconstruct previously observed content from contextual memory, reflecting its understanding of scene dynamics and physical laws.  
Given a full memory sequence \( \mathcal{M} = \{f_1, \dots, f_T\} \) and an action sequence \( \mathcal{A} = \{a_{T+1}, \dots, a_{T+k}\} \), the model generates predicted frames \( \hat{\mathcal{V}} = \{\hat{f}_{T+1}, \dots, \hat{f}_{T+k}\} \).  
Ideally, the predicted sequence should match the real sequence \( \mathcal{V} = \{f_{T+1}, \dots, f_{T+k}\} \) obtained under the same actions.  
We quantify the long-context memory ability by the mean squared error (MSE) between predicted and ground-truth frames:
$\mathcal{L}_{\text{lcm}} = \frac{1}{k} \sum_{i=1}^{k} \| \hat{f}_{T+i} - f_{T+i} \|_2^2$
where a lower \( \mathcal{L}_{\text{lcm}} \) indicates stronger long-term memory retention and reconstruction fidelity.


\textbf{Generated Scene Consistency}.
To quantify the world model's ability to maintain consistency in generated scenes, we introduce a \textit{generated scene consistency} metric based on $10$ symmetric motion paths (Figure~\ref{fig-mirror-path}), each involving simple translations or rotations lasting $24$ frames.
The model moves forward and then retraces the same path in reverse; ideally, frames from the forward (\texttt{fwd}) and reverse (\texttt{rev}) paths should match exactly.
We measure this consistency using MSE:
$\mathcal{L}_{\text{gsc}} = \frac{1}{k} \sum_{i=1}^{k} \| \hat{f}_{T+i}^{\text{fwd}} - \hat{f}_{T+i}^{\text{rev}} \|_2^2,$
where \( \hat{f}_{T+i}^{\text{fwd}} \) and \( \hat{f}_{T+i}^{\text{rev}} \) denote the predicted frames from the forward and reverse trajectories, respectively.  
A lower $\mathcal{L}_{\text{gsc}}$ indicates stronger scene generation consistency and geometric stability.

\textbf{Action Accuracy.}
The accuracy of action feedback in world models is central to their precise instruction execution and reliable completion of complex sequential tasks. To evaluate this capability fairly, we unify the predefined action sequences for all models, recover camera trajectories from generated videos via \texttt{ViPE} \cite{huang2025vipe}, eliminate scale and coordinate system discrepancies through Sim(3) Umeyama alignment \cite{umeyama2002least}, and then calculate translational and rotational relative pose errors. This metric quantifies the accuracy of each model in generating expected frames based on action commands, and is independent of the model's internal velocity parameters.

\textbf{Action Space Generalization.}
World models serve as simulators for domains such as autonomous driving and robotics, where understanding spatial regularities is crucial.
We evaluate \textit{action space generalization} by computing the MSE between generated and ground-truth frames under diverse action settings.
Ideally, the model should learn action-space constraints from context and generate videos that follow them with zero-shot consistency.

\textbf{Visual Quality.}
We evaluate visual quality from two complementary perspectives: 1) \textit{Aesthetic quality}. \texttt{LAION}~\citep{schuhmann2022laionaesthetic} aesthetic prediction model is used to quantitatively evaluate the visual attractiveness of each frame. 
Trained on large-scale human preference data, it scores composition, color, lighting, realism, and style consistency.
Higher scores indicate closer alignment with human aesthetic judgments.
2) \textit{Imaging Quality.} \texttt{MUSIQ}~\cite{ke2021musiq} evaluates perceptual fidelity by detecting artifacts like overexposure, noise, compression, and blur.
Trained on the SPAQ~\cite{fang2020perceptual} dataset, it quantifies image clarity and sharpness as an objective measure of visual quality.

\section{Experiment}
\begin{figure}[t]
    \centering
    \includegraphics[width=1\linewidth]{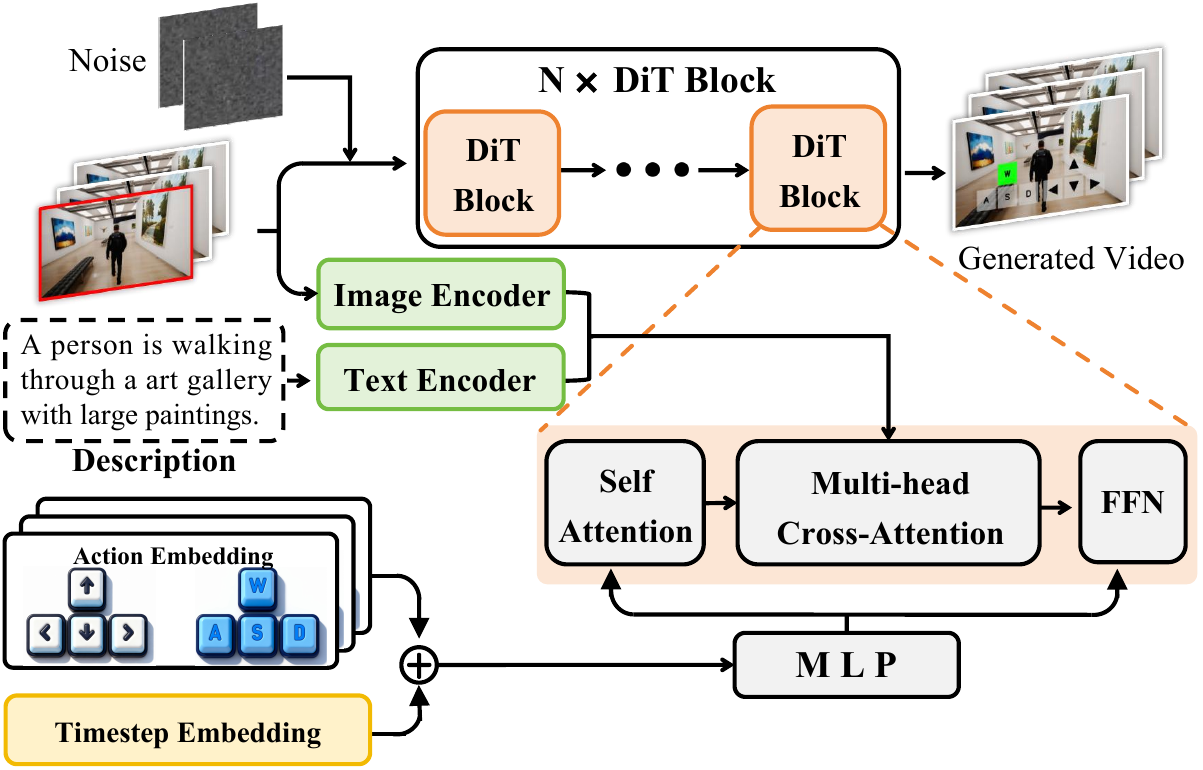}
    \vspace{-0.4cm}
    \caption{\textbf{MIND-World model framework}. 
    The parameterized action injection mechanism enables frame-level alignment with input videos and streamlines the process, forming an efficient baseline for Video-to-World training and inference. Furthermore, it allows for unlimited-length inference based on action sequences.}
    \label{fig-SkyReels-pipeline}
    \vspace{-0.4cm}
\end{figure}
\subsection{MIND-World}
We present \texttt{MIND-World}, a high-dynamics, real-time, interactive autoregressive video generation pipeline (see Figure~\ref{fig-SkyReels-pipeline}). 
Following~\cite{he2025matrix}, our training pipeline has three stages: 
(i) a bidirectional, action-conditioned teacher model,
(ii) student initialization from the teacher’s ODE trajectories~\cite{yin2025causvid}, and
(iii) self-forcing DMD distillation~\cite{huang2025self} into a few-step, frame-wise causal pipeline.
Unlike~\cite{yu2025gamefactory, he2025matrix}, which relies on heavy action blocks, we inject actions directly into the timestep embeddings, yielding a simpler and effective conditioning mechanism.
At inference time, we maintain a context cache and generate frames autoregressively conditioned on both the cached context and incoming actions, enabling continuous, low-latency streaming generation. 
%
We evaluate under two settings:
1) With context memory: a window of $w$ frames is cached as clean world context in working memory, conditioning subsequent frame generation.
2) Without context memory: generation cold-starts from the initial image and proceeds autoregressively.

\begin{table*}[t]
    \centering
      \footnotesize
      \newcommand{\best}[1]{\textbf{#1}}
      \caption{\textbf{Model Performance for the First Person on \ours-First 50}. All videos underwent identical processing and were evaluated at 720p resolution. $\downarrow$ indicates lower values are better; $\uparrow$ indicates higher values are better.}
    \vspace{-0.2cm}
    \label{tab:main_exp_first_person}
    \begin{tabular}{
  l |c| c| c| c| c| c| c
}
\toprule
\multirow{1}{*}{\textbf{Model}} &
\multicolumn{1}{c|}{\makecell{Long \\ Context Mem.}$\downarrow$} &
\multicolumn{1}{c|}{\makecell{Generated\\ Scene Consis.}$\downarrow$} &
\multicolumn{1}{c|}{\makecell{Action Space \\ Generalization}$\downarrow$}&
\multicolumn{1}{c|}{\makecell{Aesthetic \\ Quality}$\uparrow$}&
\multicolumn{1}{c|}{\makecell{Image \\ Quality}$\uparrow$}&
\multicolumn{2}{c}{\makecell{Action Accuracy (RPE$\downarrow$)\\
\!\!\!\!\!\!\!\!\!\!Trans \space\space\space\space\space\space\space\space\space\space\space \space\space\space Rot}}
\\
\midrule

\multicolumn{8}{c}{\red{\textit{w/o Context Memory (Image-to-World)}}} \\
MIND-World (Ours) & 0.1091&0.0359 & 0.1200&0.4583&0.5655 &0.0356 & \textbf{0.4395}\\
Matrix-Game 2.0 \cite{he2025matrix} & 0.1188 & \textbf{0.0306} & \textbf{0.1084} & 0.4302 &0.5180 & \textbf{0.0265}  & 0.6914\\
\midrule
\addlinespace[2pt]
\multicolumn{8}{c}{\color{green!80!black}{\textit{w Context Memory (Video-to-World)}}} \\
MIND-World (Ours) & \textbf{0.1035} & 0.0309&0.1226&\textbf{0.4590}&\textbf{0.5702}  &0.0384& 0.5534\\
\bottomrule
\end{tabular}
\end{table*}

\begin{table*}[t]
    \centering
      \footnotesize
      \newcommand{\best}[1]{\textbf{#1}}
      \caption{\textbf{Model Performance for the Third Person on \ours-Third 50}. All videos underwent identical processing and were evaluated at 720p resolution. $\downarrow$ indicates lower values are better; $\uparrow$ indicates higher values are better.}
      \vspace{-0.2cm}
    \label{tab:main_exp_third_person}
    \begin{tabular}{
  l |c| c| c| c| c| c| c
}
\toprule
\multirow{1}{*}{\textbf{Model}} &
\multicolumn{1}{c|}{\makecell{Long \\ Context Mem.}$\downarrow$} &
\multicolumn{1}{c|}{\makecell{Generated \\ Scene Consis.}$\downarrow$} &
\multicolumn{1}{c|}{\makecell{Action Space \\ Generalization}$\downarrow$}&
\multicolumn{1}{c|}{\makecell{Aesthetic \\ Quality}$\uparrow$}&
\multicolumn{1}{c|}{\makecell{Image \\ Quality}$\uparrow$}&
\multicolumn{2}{c}{\makecell{Action Accuracy (RPE$\downarrow$)\\
\!\!\!\!\!\!\!\!\!\!Trans \space\space\space\space\space\space\space\space\space\space\space \space\space\space Rot}}
\\
\midrule

\multicolumn{8}{c}{\red{\textit{w/o Context Memory (Image-to-World)}}} \\
MIND-World (Ours) &0.1066 &0.0327 &\textbf{0.0677} & 0.5204& 0.5672& \textbf{0.0271}& \textbf{0.2587}\\
Matrix-Game 2.0 \cite{he2025matrix}& 0.1404 & 0.0372 & 0.0777 & 0.4236&0.4857 & 0.0622 & 0.9031\\
\midrule
\addlinespace[2pt]
\multicolumn{8}{c}{\color{green!80!black}{\textit{w Context Memory (Video-to-World)}}} \\
MIND-World (Ours) & \textbf{0.1042} & \textbf{0.0316} & 0.0685 & \textbf{0.5300}&\textbf{0.5673}& 0.0321 &0.3338 \\
\bottomrule
\end{tabular}
\end{table*}

\subsection{Experiment Setting}
\textbf{MIND-World.} 
We initialize the foundation model with \texttt{SkyReels-V2-I2V-1.3B}~\cite{chen2025skyreelsv2infinitelengthfilmgenerative} and fine-tune it with action injection for $3$K steps with batch size $8$. For distillation, we initialize the $4$-step causal student based 1K teacher's ODE trajectories and train for $3$K steps, followed by $2$K steps via DMD-based Self-Forcing.
The student is strictly per-frame causal (chunk size = $1$) with a local attention window of \textbf{25} frames.
All experiments are conducted on $4$~$\times$~NVIDIA H100 GPUs.

\textbf{\ours Dataset.} As illustrated in Figure \ref{fig-scene}, the dataset covers $8$ major categories, comprising a total of $100$ first-person and $100$ third-person videos in the same action space, along with $25$ first-person and $25$ third-person videos in different action spaces.
All videos are long-term, open-domain, high-quality, and frame-aligned. As illustrated in Figure \ref{fig-scene_number}, during the finetune training and test split, action distribution consistency is ensured, and the distribution across scene categories is balanced as much as possible. Among them, $50$ first-person and $50$ third-person videos from the shared action space are used for training \texttt{MIND-World}, while the remaining $150$ videos are reserved for the \ours evaluation.

\subsection{Per-Dimension Evaluation}
For each dimension, we compute the \ours score using the evaluation suite described in Section \ref{Evaluation}, with results presented in Tables \ref{tab:main_exp_first_person} and \ref{tab:main_exp_third_person}. To advance future research, we introduce \texttt{MIND-World} as an open-domain video-to-world baseline with memory-augmented world modeling capabilities.

\textbf{Memory Consistency.}
Tables \ref{tab:main_exp_first_person} and \ref{tab:main_exp_third_person} show that, on the long context memory metric, models with context memory outperform those without by more than $4\%$. generated scene consistency results further confirm the benefits of memory. Additionally, \texttt{Matrix-game-2.0} performs poorly in third-person generation; human evaluation verifies that its metrics accurately reflect this limitation—the model fails to generate controllable third-person characters.

\textbf{Action Accuracy.}
As shown in Tables \ref{tab:main_exp_first_person} and \ref{tab:main_exp_third_person}, even when inputting context memory with the same action space as the fine-tuning phase, the world model's action control performance still deteriorates. This indicates limitations in the current action injection mechanism, and how to design more effective action injection strategies to enhance the world model's action control capability remains an important research problem worthy of in-depth exploration.

\textbf{Action Space Generalization.}
Tables \ref{tab:main_exp_first_person} and \ref{tab:main_exp_third_person} show that injecting context memory impairs world model inference, since inconsistent action spaces disrupt reasoning in models without action generalization.

\textbf{Visual Quality.}
Tables \ref{tab:main_exp_first_person} and \ref{tab:main_exp_third_person} collectively show that world models with memory produce videos of superior visual quality and better alignment with human aesthetic preferences. This is due to the memory mechanism leveraging richer contextual information, ensuring high consistency in style and coherence with the given segment. 


\begin{figure*}[t]
    \centering
    \includegraphics[width=0.99\linewidth]{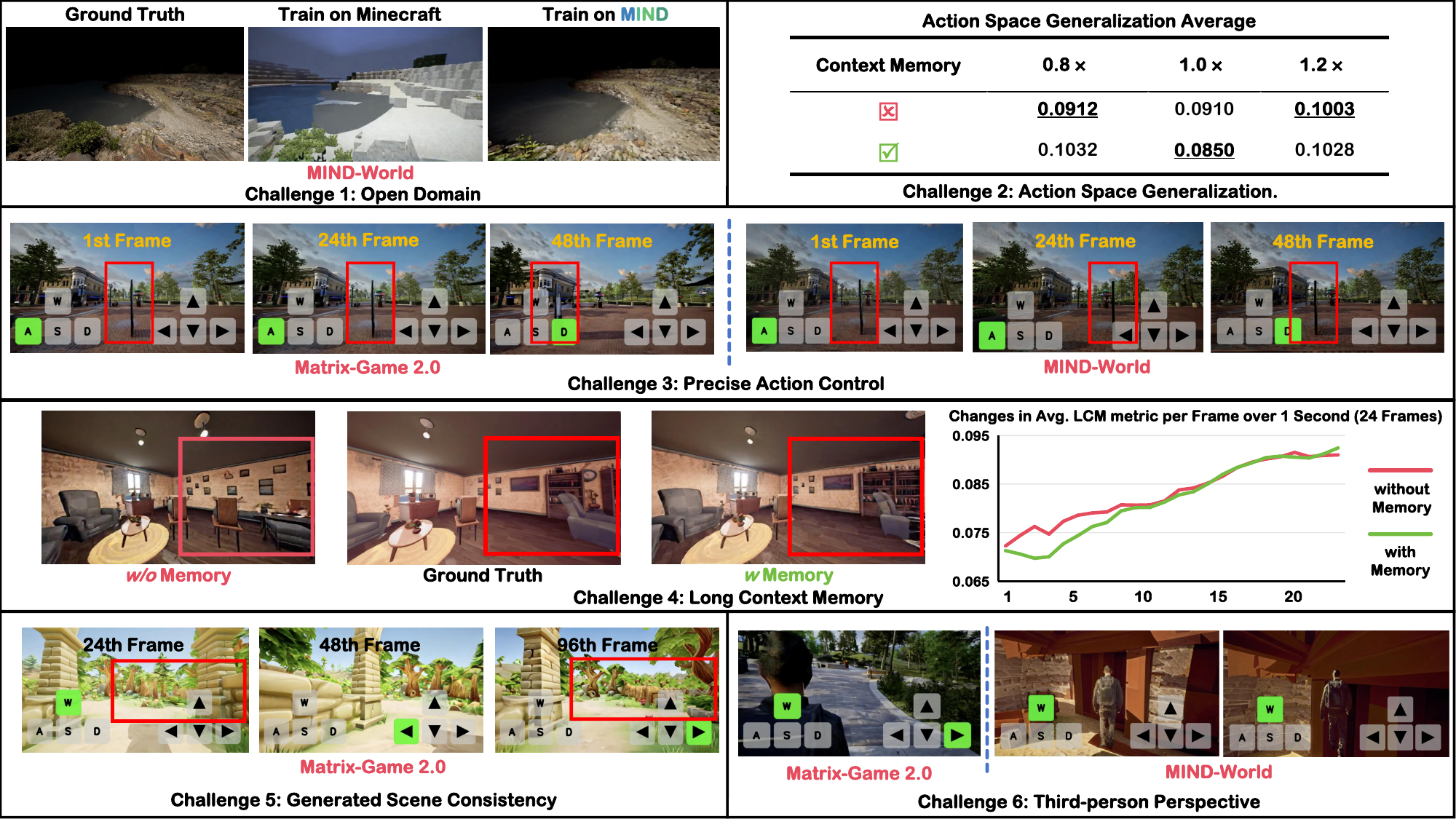}
    \vspace{-0.2cm}
    \caption{\textbf{Summary of insights from the challenges in \ours}. 
    For each challenge, a representative example is visualized. }
    \label{fig-challenge}
\end{figure*}
\subsection{Insights and Discussions}
This section details the observations and insights derived from our comprehensive evaluation experiments.

\textbf{Open Domain.}
As illustrated in Challenge $1$ of Figure \ref{fig-challenge},    
\texttt{MIND-World} trained on easily collected Minecraft datasets struggle to generalize to open-domain inference, whereas those trained on the high-quality dataset provided by \ours exhibit significantly improved generalization. However, acquiring such data is challenging; thus, effectively leveraging readily available large-scale data to achieve open-domain generalization remains a key open problem.

\textbf{Action-Space Generalization.}
Tables \ref{tab:main_exp_first_person} and \ref{tab:main_exp_third_person} reveal that, in action space generalization, video-to-world models with memory capabilities underperform image-to-world models without memory. As shown in Challenge $2$ of Figure \ref{fig-challenge}, further analysis indicates that memory-enabled models outperform memory-less ones within the original action space; however, their performance drops significantly when the action space changes. This suggests that context memory tied to an action space inconsistent with training disrupts model inference. Therefore, accurately detecting the action space from context memory and achieving precise prediction remains a major challenge.

\textbf{Precise Action Control.}
In the experiment of Path $5$ in Figure \ref{fig-mirror-path}, where the agent first moves left and then right to return to the starting position—the generated result is shown in Challenge 3 of Figure~\ref{fig-challenge}, \texttt{Matrix-game-2.0} \cite{he2025matrix} fails to move left as expected, instead remaining stationary and ultimately stopping far to the right of the origin.
In contrast, \texttt{MIND-World} correctly moves left but fails to return to the initial position after moving right.
Repeated experiments reveal that the visual prompt (\textit{i.e.,} the input image or video) significantly affects action following.
Therefore, separating visual prompts from action dynamics is key to achieving precise action control in world models.

\textbf{Long-horizon Memory Consistency.}
Tables \ref{tab:main_exp_first_person} and \ref{tab:main_exp_third_person} show that, in long-horizon rollouts, models with memory significantly outperform those without. The visualization in Challenge $4$ of Figure \ref{fig-challenge} further confirms this: memory-enabled generations remain largely consistent with ground truth, while memory-less outputs deviate substantially. Moreover, current world models can only capture short-term memory; effectively maintaining and leveraging long-context memory remains a critical open problem.

\textbf{Generated Scene Consistency.}
By conducting a mirroring experiment on \texttt{Matrix-game-2.0} \cite{he2025matrix} along Path $5$ in Figure \ref{fig-mirror-path}, the results as shown in Challenge $5$ of Figure \ref{fig-challenge} reveal that when the camera revisits previously generated scenes, the content is clearly inconsistent with prior generations. Therefore, ensuring consistent generation of scenes continues to pose a significant difficulty.

\textbf{Third-person Perspective.}
As shown in Challenge $6$ of Figure \ref{fig-challenge}, \texttt{Matrix-game-2.0} \cite{he2025matrix} fails to control the third-person character and execute movement, causing the generated video to pass through the character and eventually lose it. In contrast, \texttt{MIND-World} can control the character but fails to properly handle the relationship between the foreground character and the background, resulting in the character passing directly through buildings. Therefore, accurately perceiving and modeling the interactions between characters and backgrounds remains a major challenge in world models.
\section{Conclusion}
\label{sec:Conclusion}

We introduced \ours, the first open-domain closed-loop revisited benchmark for evaluating memory consistency and action control in world models from both first-person and third-person perspectives.
Built on Unreal Engine 5 with diverse action spaces, \ours enables systematic assessment of long-term scene memory, temporal coherence, and action space generalization. 
Experiments with \texttt{MIND-World} reveal that the challenges remain in generalizing across action spaces and maintaining long-horizon coherence. 
\ours establishes a unified foundation for advancing interactive, temporally consistent open domain world model.

\bibliographystyle{ieeenat_fullname}
\bibliography{main}

@String(CVPR= {IEEE Conf. Comput. Vis. Pattern Recog.})

@String(ICCV= {Int. Conf. Comput. Vis.})

@String(CVPR  = {CVPR})

@String(ICCV  = {ICCV})

@article{valevski2024diffusion,
  title={Diffusion models are real-time game engines},
  author={Valevski, Dani and Leviathan, Yaniv and Arar, Moab and Fruchter, Shlomi},
  journal={arXiv preprint arXiv:2408.14837},
  year={2024}
}

@article{che2024gamegen,
  title={Gamegen-x: Interactive open-world game video generation},
  author={Che, Haoxuan and He, Xuanhua and Liu, Quande and Jin, Cheng and Chen, Hao},
  journal={arXiv preprint arXiv:2411.00769},
  year={2024}
}

@article{he2025matrix,
    title={Matrix-Game 2.0: An Open-Source, Real-Time, and Streaming Interactive World Model},
    author={He, Xianglong and Peng, Chunli and Liu, Zexiang and Wang, Boyang and Zhang, Yifan and Cui, Qi and Kang, Fei and Jiang, Biao and An, Mengyin and Ren, Yangyang and Xu, Baixin and Guo, Hao-Xiang and Gong, Kaixiong and Wu, Cyrus and Li, Wei and Song, Xuchen and Liu, Yang and Li, Eric and Zhou, Yahui},
    journal={arXiv preprint arXiv:2508.13009},
    year={2025}
}

@article{yu2025gamefactory,
  title={Gamefactory: Creating new games with generative interactive videos},
  author={Yu, Jiwen and Qin, Yiran and Wang, Xintao and Wan, Pengfei and Zhang, Di and Liu, Xihui},
  journal={arXiv preprint arXiv:2501.08325},
  year={2025}
}

@article{mao2025yume,
  title={Yume: An Interactive World Generation Model},
  author={Mao, Xiaofeng and Lin, Shaoheng and Li, Zhen and Li, Chuanhao and Peng, Wenshuo and He, Tong and Pang, Jiangmiao and Chi, Mingmin and Qiao, Yu and Zhang, Kaipeng},
  journal={arXiv preprint arXiv:2507.17744},
  year={2025}
}

@article{zhang2025matrix,
  title={Matrix-Game: Interactive World Foundation Model},
  author={Zhang, Yifan and Peng, Chunli and Wang, Boyang and Wang, Puyi and Zhu, Qingcheng and Kang, Fei and Jiang, Biao and Gao, Zedong and Li, Eric and Liu, Yang and others},
  journal={arXiv preprint arXiv:2506.18701},
  year={2025}
}

@inproceedings{fang2020perceptual,
  title={Perceptual quality assessment of smartphone photography},
  author={Fang, Yuming and Zhu, Hanwei and Zeng, Yan and Ma, Kede and Wang, Zhou},
  booktitle={Proceedings of the IEEE/CVF conference on computer vision and pattern recognition},
  pages={3677--3686},
  year={2020}
}

@inproceedings{ke2021musiq,
  title={Musiq: Multi-scale image quality transformer},
  author={Ke, Junjie and Wang, Qifei and Wang, Yilin and Milanfar, Peyman and Yang, Feng},
  booktitle={Proceedings of the IEEE/CVF international conference on computer vision},
  pages={5148--5157},
  year={2021}
}

@article{ye2025yan,
  title={Yan: Foundational interactive video generation},
  author={Ye, Deheng and Zhou, Fangyun and Lv, Jiacheng and Ma, Jianqi and Zhang, Jun and Lv, Junyan and Li, Junyou and Deng, Minwen and Yang, Mingyu and Fu, Qiang and others},
  journal={arXiv preprint arXiv:2508.08601},
  year={2025}
}

@article{yang2025raw2drive,
  title={Raw2Drive: Reinforcement learning with aligned world models for end-to-end autonomous driving (in carla v2)},
  author={Yang, Zhenjie and Jia, Xiaosong and Li, Qifeng and Yang, Xue and Yao, Maoqing and Yan, Junchi},
  journal={arXiv preprint arXiv:2505.16394},
  year={2025}
}

@article{mousakhan2025orbis,
  title={Orbis: Overcoming challenges of long-horizon prediction in driving world models},
  author={Mousakhan, Arian and Mittal, Sudhanshu and Galesso, Silvio and Farid, Karim and Brox, Thomas},
  journal={arXiv preprint arXiv:2507.13162},
  year={2025}
}

@article{li2025drivevla,
  title={DriveVLA-W0: World Models Amplify Data Scaling Law in Autonomous Driving},
  author={Li, Yingyan and Shang, Shuyao and Liu, Weisong and Zhan, Bing and Wang, Haochen and Wang, Yuqi and Chen, Yuntao and Wang, Xiaoman and An, Yasong and Tang, Chufeng and others},
  journal={arXiv preprint arXiv:2510.12796},
  year={2025}
}

@article{cen2025worldvla,
  title={WorldVLA: Towards Autoregressive Action World Model},
  author={Cen, Jun and Yu, Chaohui and Yuan, Hangjie and Jiang, Yuming and Huang, Siteng and Guo, Jiayan and Li, Xin and Song, Yibing and Luo, Hao and Wang, Fan and others},
  journal={arXiv preprint arXiv:2506.21539},
  year={2025}
}

@article{lv2025f1,
  title={F1: A vision-language-action model bridging understanding and generation to actions},
  author={Lv, Qi and Kong, Weijie and Li, Hao and Zeng, Jia and Qiu, Zherui and Qu, Delin and Song, Haoming and Chen, Qizhi and Deng, Xiang and Pang, Jiangmiao},
  journal={arXiv preprint arXiv:2509.06951},
  year={2025}
}

@inproceedings{bruce2024genie,
  title={Genie: Generative interactive environments},
  author={Bruce, Jake and Dennis, Michael D and Edwards, Ashley and Parker-Holder, Jack and Shi, Yuge and Hughes, Edward and Lai, Matthew and Mavalankar, Aditi and Steigerwald, Richie and Apps, Chris and others},
  booktitle={ICML},
  year={2024}
}

@misc{schuhmann2022laionaesthetic,
  title        = {LAION-Aesthetic Predictor},
  author       = {Christoph Schuhmann and Richard Vencu and Romain Beaumont and Ross Wightman and Mitchell Wortsman and Mehdi Cherti and Clayton Mullis and Andreas Köpf and Theo Coombes and Jenia Jitsev},
  year         = {2022},
  howpublished = {\url{https://github.com/LAION-AI/aesthetic-predictor}},
  note         = {LAION-AI, GitHub repository}
}

@article{duan2025worldscore,
    title={WorldScore: A Unified Evaluation Benchmark for World Generation},
    author={Duan, Haoyi and Yu, Hong-Xing and Chen, Sirui and Fei-Fei, Li and Wu, Jiajun},
    journal={ICCV},
    year={2025}
}

@article{qin2025worldsimbench,
  title={WorldSimBench: Towards Video Generation Models as World Simulators},
  author={Qin, Yiran and Shi, Zhelun and Yu, Jiwen and Wang, Xijun and Zhou, Enshen and Li, Lijun and Yin, Zhenfei and Liu, Xihui and Sheng, Lu and Shao, Jing and others},
  journal={ICML},
  year={2025}
}

@article{lian2025toward,
  title={Toward Memory-Aided World Models: Benchmarking via Spatial Consistency},
  author={Lian, Kewei and Cai, Shaofei and Du, Yilun and Liang, Yitao},
  journal={arXiv preprint arXiv:2505.22976},
  year={2025}
}

@inproceedings{Li2025WorldModelBench,
            title={WorldModelBench: Judging Video Generation Models As World Models},
            author={Dacheng Li and Yunhao Fang and Yukang Chen and Shuo Yang and Shiyi Cao and Justin Wong and Michael Luo and Xiaolong Wang and Hongxu Yin and Joseph E. Gonzalez and Ion Stoica and Song Han and Yao Lu},
            year={2025},
          }

@article{zhang2025world,
  title={World-in-World: World Models in a Closed-Loop World},
  author={Zhang, Jiahan and Jiang, Muqing and Dai, Nanru and Lu, Taiming and Uzunoglu, Arda and Zhang, Shunchi and Wei, Yana and Wang, Jiahao and Patel, Vishal M and Liang, Paul Pu and others},
  journal={arXiv preprint arXiv:2510.18135},
  year={2025}
}

@article{svd,
  title={Stable video diffusion: Scaling latent video diffusion models to large datasets},
  author={Blattmann, Andreas and Dockhorn, Tim and Kulal, Sumith and Mendelevitch, Daniel and Kilian, Maciej and Lorenz, Dominik and Levi, Yam and English, Zion and Voleti, Vikram and Letts, Adam and others},
  journal={arXiv preprint arXiv:2311.15127},
  year={2023}
}

@inproceedings{huang2024vbench,
  title={Vbench: Comprehensive benchmark suite for video generative models},
  author={Huang, Ziqi and He, Yinan and Yu, Jiashuo and Zhang, Fan and Si, Chenyang and Jiang, Yuming and Zhang, Yuanhan and Wu, Tianxing and Jin, Qingyang and Chanpaisit, Nattapol and others},
  booktitle={Proceedings of the IEEE/CVF Conference on Computer Vision and Pattern Recognition},
  pages={21807--21818},
  year={2024}
}

@article{zheng2025vbench,
  title={Vbench-2.0: Advancing video generation benchmark suite for intrinsic faithfulness},
  author={Zheng, Dian and Huang, Ziqi and Liu, Hongbo and Zou, Kai and He, Yinan and Zhang, Fan and Gu, Lulu and Zhang, Yuanhan and He, Jingwen and Zheng, Wei-Shi and others},
  journal={arXiv preprint arXiv:2503.21755},
  year={2025}
}

@article{chen2024diffusion,
  title={Diffusion forcing: Next-token prediction meets full-sequence diffusion},
  author={Chen, Boyuan and Mart{\'\i} Mons{\'o}, Diego and Du, Yilun and Simchowitz, Max and Tedrake, Russ and Sitzmann, Vincent},
  journal={Advances in Neural Information Processing Systems},
  volume={37},
  pages={24081--24125},
  year={2024}
}

@article{huang2025self,
  title={Self Forcing: Bridging the Train-Test Gap in Autoregressive Video Diffusion},
  author={Huang, Xun and Li, Zhengqi and He, Guande and Zhou, Mingyuan and Shechtman, Eli},
  journal={arXiv preprint arXiv:2506.08009},
  year={2025}
}

@misc{openai_sora2_2025,
  author       = {OpenAI},
  title        = {Sora 2 is here: our latest video generation model},
  howpublished = {\url{https://openai.com/index/sora-2/}},
  month        = {September},
  year         = {2025},
  note         = {Accessed: 2025-10-29}
}

@article{kong2024hunyuanvideo,
  title={Hunyuanvideo: A systematic framework for large video generative models},
  author={Kong, Weijie and Tian, Qi and Zhang, Zijian and Min, Rox and Dai, Zuozhuo and Zhou, Jin and Xiong, Jiangfeng and Li, Xin and Wu, Bo and Zhang, Jianwei and others},
  journal={arXiv preprint arXiv:2412.03603},
  year={2024}
}

@article{wan2025wan,
  title={Wan: Open and advanced large-scale video generative models},
  author={Wan, Team and Wang, Ang and Ai, Baole and Wen, Bin and Mao, Chaojie and Xie, Chen-Wei and Chen, Di and Yu, Feiwu and Zhao, Haiming and Yang, Jianxiao and others},
  journal={arXiv preprint arXiv:2503.20314},
  year={2025}
}

@misc{chen2025skyreelsv2infinitelengthfilmgenerative,
      title={{SkyReels-V2}: Infinite-length Film Generative Model}, 
      author={Guibin Chen and Dixuan Lin and Jiangping Yang and Chunze Lin and Junchen Zhu and Mingyuan Fan and Hao Zhang and Sheng Chen and Zheng Chen and Chengcheng Ma and Weiming Xiong and Wei Wang and Nuo Pang and Kang Kang and Zhiheng Xu and Yuzhe Jin and Yupeng Liang and Yubing Song and Peng Zhao and Boyuan Xu and Di Qiu and Debang Li and Zhengcong Fei and Yang Li and Yahui Zhou},
      year={2025},
      eprint={2504.13074},
      archivePrefix={arXiv},
      primaryClass={cs.CV},
      url={https://arxiv.org/abs/2504.13074}, 
}

@inproceedings{yin2025causvid,
    title={From Slow Bidirectional to Fast Autoregressive Video Diffusion Models},
    author={Yin, Tianwei and Zhang, Qiang and Zhang, Richard and Freeman, William T and Durand, Fredo and Shechtman, Eli and Huang, Xun},
    booktitle={CVPR},
    year={2025}
}

@misc{panteam2025panworldmodelgeneral,
      title={PAN: A World Model for General, Interactable, and Long-Horizon World Simulation}, 
      author={PAN Team and Jiannan Xiang and Yi Gu and Zihan Liu and Zeyu Feng and Qiyue Gao and Yiyan Hu and Benhao Huang and Guangyi Liu and Yichi Yang and Kun Zhou and Davit Abrahamyan and Arif Ahmad and Ganesh Bannur and Junrong Chen and Kimi Chen and Mingkai Deng and Ruobing Han and Xinqi Huang and Haoqiang Kang and Zheqi Li and Enze Ma and Hector Ren and Yashowardhan Shinde and Rohan Shingre and Ramsundar Tanikella and Kaiming Tao and Dequan Yang and Xinle Yu and Cong Zeng and Binglin Zhou and Zhengzhong Liu and Zhiting Hu and Eric P. Xing},
      year={2025},
      eprint={2511.09057},
      archivePrefix={arXiv},
      primaryClass={cs.CV},
      url={https://arxiv.org/abs/2511.09057}, 
}

@misc{li2025hunyuangamecrafthighdynamicinteractivegame,
      title={Hunyuan-GameCraft: High-dynamic Interactive Game Video Generation with Hybrid History Condition}, 
      author={Jiaqi Li and Junshu Tang and Zhiyong Xu and Longhuang Wu and Yuan Zhou and Shuai Shao and Tianbao Yu and Zhiguo Cao and Qinglin Lu},
      year={2025},
      eprint={2506.17201},
      archivePrefix={arXiv},
      primaryClass={cs.CV},
      url={https://arxiv.org/abs/2506.17201}, 
}

@misc{li2025omninwmomniscientdrivingnavigation,
      title={OmniNWM: Omniscient Driving Navigation World Models}, 
      author={Bohan Li and Zhuang Ma and Dalong Du and Baorui Peng and Zhujin Liang and Zhenqiang Liu and Chao Ma and Yueming Jin and Hao Zhao and Wenjun Zeng and Xin Jin},
      year={2025},
      eprint={2510.18313},
      archivePrefix={arXiv},
      primaryClass={cs.CV},
      url={https://arxiv.org/abs/2510.18313}, 
}

@misc{cui2025emu35nativemultimodalmodels,
      title={Emu3.5: Native Multimodal Models are World Learners}, 
      author={Yufeng Cui and Honghao Chen and Haoge Deng and Xu Huang and Xinghang Li and Jirong Liu and Yang Liu and Zhuoyan Luo and Jinsheng Wang and Wenxuan Wang and Yueze Wang and Chengyuan Wang and Fan Zhang and Yingli Zhao and Ting Pan and Xianduo Li and Zecheng Hao and Wenxuan Ma and Zhuo Chen and Yulong Ao and Tiejun Huang and Zhongyuan Wang and Xinlong Wang},
      year={2025},
      eprint={2510.26583},
      archivePrefix={arXiv},
      primaryClass={cs.CV},
      url={https://arxiv.org/abs/2510.26583}, 
}

@misc{zhang2025epona,
      title={Epona: Autoregressive Diffusion World Model for Autonomous Driving}, 
      author={Kaiwen Zhang and Zhenyu Tang and Xiaotao Hu and Xingang Pan and Xiaoyang Guo and Yuan Liu and Jingwei Huang and Li Yuan and Qian Zhang and Xiao-Xiao Long and Xun Cao and Wei Yin},
      year={2025},
      eprint={2506.24113},
      archivePrefix={arXiv},
      primaryClass={cs.CV},
      url={https://arxiv.org/abs/2506.24113}, 
}

@misc{chi2025wow,
      title={WoW: Towards a World omniscient World model Through Embodied Interaction}, 
      author={Xiaowei Chi and Peidong Jia and Chun-Kai Fan and Xiaozhu Ju and Weishi Mi and Kevin Zhang and Zhiyuan Qin and Wanxin Tian and Kuangzhi Ge and Hao Li and Zezhong Qian and Anthony Chen and Qiang Zhou and Yueru Jia and Jiaming Liu and Yong Dai and Qingpo Wuwu and Chengyu Bai and Yu-Kai Wang and Ying Li and Lizhang Chen and Yong Bao and Zhiyuan Jiang and Jiacheng Zhu and Kai Tang and Ruichuan An and Yulin Luo and Qiuxuan Feng and Siyuan Zhou and Chi-min Chan and Chengkai Hou and Wei Xue and Sirui Han and Yike Guo and Shanghang Zhang and Jian Tang},
      year={2025},
      eprint={2509.22642},
      archivePrefix={arXiv},
      primaryClass={cs.RO},
      url={https://arxiv.org/abs/2509.22642}, 
}

@misc{xiao2025worldmem,
      title={WORLDMEM: Long-term Consistent World Simulation with Memory}, 
      author={Zeqi Xiao and Yushi Lan and Yifan Zhou and Wenqi Ouyang and Shuai Yang and Yanhong Zeng and Xingang Pan},
      year={2025},
      eprint={2504.12369},
      archivePrefix={arXiv},
      primaryClass={cs.CV},
      url={https://arxiv.org/abs/2504.12369}, 
}

@misc{wu2025videoworldmodelslongterm,
      title={Video World Models with Long-term Spatial Memory}, 
      author={Tong Wu and Shuai Yang and Ryan Po and Yinghao Xu and Ziwei Liu and Dahua Lin and Gordon Wetzstein},
      year={2025},
      eprint={2506.05284},
      archivePrefix={arXiv},
      primaryClass={cs.CV},
      url={https://arxiv.org/abs/2506.05284}, 
}

@misc{huang2025memoryforcing,
      title={Memory Forcing: Spatio-Temporal Memory for Consistent Scene Generation on Minecraft}, 
      author={Junchao Huang and Xinting Hu and Boyao Han and Shaoshuai Shi and Zhuotao Tian and Tianyu He and Li Jiang},
      year={2025},
      eprint={2510.03198},
      archivePrefix={arXiv},
      primaryClass={cs.CV},
      url={https://arxiv.org/abs/2510.03198}, 
}

@misc{li2025vmem,
      title={VMem: Consistent Interactive Video Scene Generation with Surfel-Indexed View Memory}, 
      author={Runjia Li and Philip Torr and Andrea Vedaldi and Tomas Jakab},
      year={2025},
      eprint={2506.18903},
      archivePrefix={arXiv},
      primaryClass={cs.CV},
      url={https://arxiv.org/abs/2506.18903}, 
}

@misc{yu2025contextmemory,
      title={Context as Memory: Scene-Consistent Interactive Long Video Generation with Memory Retrieval}, 
      author={Jiwen Yu and Jianhong Bai and Yiran Qin and Quande Liu and Xintao Wang and Pengfei Wan and Di Zhang and Xihui Liu},
      year={2025},
      eprint={2506.03141},
      archivePrefix={arXiv},
      primaryClass={cs.CV},
      url={https://arxiv.org/abs/2506.03141}, 
}

@article{huang2025vipe,
  title={Vipe: Video pose engine for 3d geometric perception},
  author={Huang, Jiahui and Zhou, Qunjie and Rabeti, Hesam and Korovko, Aleksandr and Ling, Huan and Ren, Xuanchi and Shen, Tianchang and Gao, Jun and Slepichev, Dmitry and Lin, Chen-Hsuan and others},
  journal={arXiv preprint arXiv:2508.10934},
  year={2025}
}

@article{umeyama2002least,
  title={Least-squares estimation of transformation parameters between two point patterns},
  author={Umeyama, Shinji},
  journal={IEEE Transactions on pattern analysis and machine intelligence},
  volume={13},
  number={4},
  pages={376--380},
  year={2002},
  publisher={IEEE}
}

@article{tang2025hunyuan,
  title={Hunyuan-GameCraft-2: Instruction-following Interactive Game World Model},
  author={Tang, Junshu and Liu, Jiacheng and Li, Jiaqi and Wu, Longhuang and Yang, Haoyu and Zhao, Penghao and Gong, Siruis and Yuan, Xiang and Shao, Shuai and Lu, Qinglin},
  journal={arXiv preprint arXiv:2511.23429},
  year={2025}
}

@article{team2026advancing,
  title={Advancing Open-source World Models},
  author={Team, Robbyant and Gao, Zelin and Wang, Qiuyu and Zeng, Yanhong and Zhu, Jiapeng and Cheng, Ka Leong and Li, Yixuan and Wang, Hanlin and Xu, Yinghao and Ma, Shuailei and others},
  journal={arXiv preprint arXiv:2601.20540},
  year={2026}
}

@article{hong2025relic,
  title={RELIC: Interactive Video World Model with Long-Horizon Memory},
  author={Hong, Yicong and Mei, Yiqun and Ge, Chongjian and Xu, Yiran and Zhou, Yang and Bi, Sai and Hold-Geoffroy, Yannick and Roberts, Mike and Fisher, Matthew and Shechtman, Eli and others},
  journal={arXiv preprint arXiv:2512.04040},
  year={2025}
}

@article{sun2025worldplay,
  title={WorldPlay: Towards Long-Term Geometric Consistency for Real-Time Interactive World Modeling},
  author={Sun, Wenqiang and Zhang, Haiyu and Wang, Haoyuan and Wu, Junta and Wang, Zehan and Wang, Zhenwei and Wang, Yunhong and Zhang, Jun and Wang, Tengfei and Guo, Chunchao},
  journal={arXiv preprint arXiv:2512.14614},
  year={2025}
}

@misc{wu2026infiniteworld,
      title={Infinite-World: Scaling Interactive World Models to 1000-Frame Horizons via Pose-Free Hierarchical Memory}, 
      author={Ruiqi Wu and Xuanhua He and Meng Cheng and Tianyu Yang and Yong Zhang and Zhuoliang Kang and Xunliang Cai and Xiaoming Wei and Chunle Guo and Chongyi Li and Ming-Ming Cheng},
      year={2026},
      eprint={2602.02393},
      archivePrefix={arXiv},
      primaryClass={cs.CV},
      url={https://arxiv.org/abs/2602.02393}, 
}

@misc{gu2025long,
      title={Long-Context Autoregressive Video Modeling with Next-Frame Prediction}, 
      author={Yuchao Gu and Weijia Mao and Mike Zheng Shou},
      year={2025},
      eprint={2503.19325},
      archivePrefix={arXiv},
      primaryClass={cs.CV},
      url={https://arxiv.org/abs/2503.19325}, 
}

@misc{zhao2025spatia,
      title={Spatia: Video Generation with Updatable Spatial Memory}, 
      author={Jinjing Zhao and Fangyun Wei and Zhening Liu and Hongyang Zhang and Chang Xu and Yan Lu},
      year={2025},
      eprint={2512.15716},
      archivePrefix={arXiv},
      primaryClass={cs.CV},
      url={https://arxiv.org/abs/2512.15716}, 
}

@misc{gao2025adaworld,
      title={AdaWorld: Learning Adaptable World Models with Latent Actions}, 
      author={Shenyuan Gao and Siyuan Zhou and Yilun Du and Jun Zhang and Chuang Gan},
      year={2025},
      eprint={2503.18938},
      archivePrefix={arXiv},
      primaryClass={cs.AI},
      url={https://arxiv.org/abs/2503.18938}, 
}



\end{document}